\documentclass[letterpaper,10pt,conference]{ieeeconf}  

\IEEEoverridecommandlockouts                              

\usepackage{graphicx} 
\usepackage{epsfig} 
\usepackage{color}
\usepackage{amsmath} 
\usepackage{amssymb}  
\usepackage{enumerate}
\usepackage{algorithm,algorithmic}
\usepackage{eso-pic}

\def\dth{\mbox{$\dot{\theta}$}}

\def\bh{\mbox{\boldmath$h$}}

\def\bq{\mbox{\boldmath$q$}}

\def\bJ{\mbox{\boldmath$J$}}

\def\bM{\mbox{\boldmath$M$}}

\def\bI{\mbox{\boldmath$I$}}

\def\blambda{\mbox{\boldmath$\lambda$}}

\def\bO{\mbox{\boldmath$O$}}

\def\mR{\mathbb{R}}

\title{\LARGE \bf
Modeling, Analysis and Activation of Planar Viscoelastically-combined
Rimless Wheels
}

\author{Fumihiko Asano, Yuxuan Xiang, Yanqiu Zheng and Cong Yan
\thanks{
The authors are with the School of Information Science, Japan Advanced
Institute of Science and Technology, 1-1 Asahidai, Nomi, Ishikawa
923-1292, Japan
        {\tt\footnotesize
\{fasano,s2110073,s2020415,gansou315\}@jaist.ac.jp}}%
}

\AddToShipoutPictureFG*{
\AtPageLowerLeft{
\ifnum\value{page}=1
\put(10,10){
\parbox{0.9\paperwidth}{
\tiny
\textcopyright\ 2022 IEEE. This is a corrected version of the IROS 2022 paper; a typographical error in Eq.~(14) has been corrected.
}}
\fi
}
}

\begin{document}

\maketitle
\thispagestyle{empty}
\pagestyle{empty}

\begin{abstract}
This paper proposes novel passive-dynamic walkers formed by two
cross-shaped frames and eight viscoelastic elements. Since it is a
combination of two four-legged rimless wheels via viscoelastic elements,
we call it viscoelastically-combined rimless wheel (VCRW). Two types of
VCRWs consisting of different cross-shaped frames are introduced; one is
formed by combining two Greek-cross-shaped frames (VCRW1), and the other
is formed by combining two-link cross-shaped frames that can rotate
freely around the central axis (VCRW2). First, we describe the model
assumptions and equations of motion and collision. Second, we
numerically analyze the basic gait properties of passive dynamic
walking. Furthermore, we consider an activation of VCRW2 for generating
a stable level gait, and discuss the significance of the study as a
novel walking support device.

\end{abstract}
\setlength{\arraycolsep}{1.2pt}
\section{INTRODUCTION}


In this paper, we propose a novel legged locomotion robot that has both
the flexibility due to the tensegrity structure
\cite{TRO,ICRA2015,Hirai} and the high stability due to the rimless
wheel (RW) \cite{Coleman,McGeer}. As a basic model, we introduce a
RW-like model in which four rigid frames are connected by eight
viscoelastic elements originally studied in \cite{CLAWAR2021}. Our
previous studies numerically showed that if the spring tension is
sufficient and the effect of the damper can be added appropriately, 
stable passive-dynamic walking can be continued without floating off the
floor during motion \cite{CLAWAR2021,Zheng2,Zheng3}. In this paper, we
introduce a new model in which two of the four rigid frames are
constrained at the center position of the frame through a common axis, 
and investigate their motion characteristics through numerical
simulations. The new model consists of two cross-shaped frames, but if
the spring tension is sufficient, the overall shape is similar to that
of an eight-legged symmetrical RW, and generating an asymptotically
stable, period-1 passive-dynamic gait is also possible. The authors
consider that this model has two novelties; one is that it combined two
side-by-side RWs unlike the conventional combined rimless wheel (CRW)
that combined the fore and rear RWs \cite{Zhao,MUBO,MIT}, and the other
is that it combined not by using a body link but by using viscoelastic
elements. Based on these features, we call the new model
viscoelastically-combined rimless wheel (VCRW). In the first half of
this paper, we introduce two different VCRW models with the same body
structure but different constraint conditions and develop the
mathematical models. We then conduct numerically simulations to show the
tendency of changes in motion characteristics with respect to the
slope. In the latter half, we introduced an active walking model in
which an upper body is added to one cross-shaped frame and a control
torque can be applied between the upper body and the frame, and generate
a stable limit-cycle gait on a horizontal plane. We also discuss its
usefulness as a walking support device.

\section{MODELING}

\subsection{Equation of Motion and Constraint Conditions}

\begin{figure*}[t]
\centering
\vspace*{-22mm}
\includegraphics[width=0.98\linewidth]{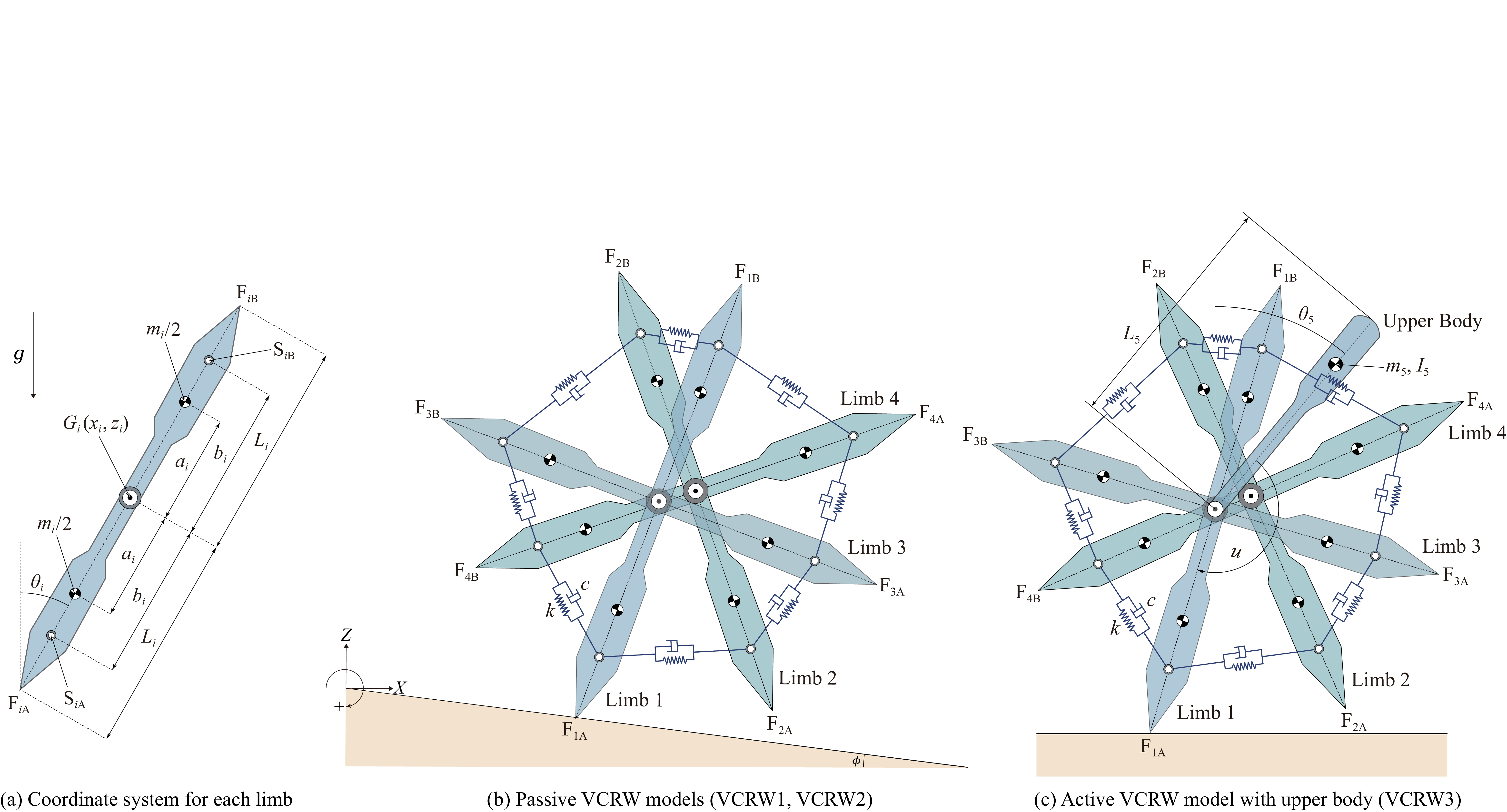}\\
\caption{Passive-dynamic and limit-cycle walker models formed by two
 cross-shaped frames and eight viscoelastic elements}
\label{fig2.01}
\end{figure*}

Fig. \ref{fig2.01} illustrates walking models. Here, (a) is the
coordinate system of each rigid frame, (b) the passive dynamic walker on
a gentle downhill whose angle is $\phi$, and (c) the active dynamic
walker with an upper body on a horizontal plane. As shown in
Fig. \ref{fig2.01}(a), let the position of one tip of Limb $i$ be ${\rm
F}_{i{\rm A}}$, and the position of the tip on the opposite side be
${\rm F}_{i{\rm B}}$. Also, let the coordinates of these tip positions
be $(x_{{\rm f}i{\rm a}}, z_{{\rm f}i{\rm a}})$ and $(x_{{\rm f}i{\rm
b}}, z_{{\rm f}i{\rm b}})$. It is assumed that the mass of each frame is
$m_i$, and its center of mass (COM) is distributed so as to be located
at the central point, $G_i$. The inertia moment around $G_i$ therefore
becomes $m_i a_i^2$ where $a_i$ is the radius of gyration. It is also
assumed that the viscoelastic element whose spring and damper
coefficients are $k$ and $c$ is attached at a point separated from $G_i$
by $b_i$. If the spring tension is sufficient, all rigid frames will
continue to rotate clockwise while maintaining a shape close to a
symmetrical eight-legged RW. Therefore, the stance foot will be
transition in the order of ${\rm F}_{1{\rm A}} \rightarrow {\rm
F}_{2{\rm A}} \rightarrow {\rm F}_{3{\rm A}} \rightarrow {\rm F}_{4{\rm
A}} \rightarrow  {\rm F}_{1{\rm B}} \rightarrow \cdots \rightarrow {\rm
F}_{4{\rm B}} \rightarrow {\rm F}_{1{\rm A}}$. 

Let $
\bq = \left[
\begin{array}{cccccccccccc}
x_1 & z_1 & \theta_1 & x_2 & z_2 & \theta_2 & x_3 & z_3 & \theta_3 & x_4
 & z_4 & \theta_4
\end{array}
\right]^{\rm T}
$ be the generalized coordinate vector. Let $P_g$ be the sum of the
potential energy due to gravity, $P_e$ be that due to elasticity, and
$R$ be the sum of the dissipative functions due to viscosity. The robot
equation of motion and holonomic constraint condition then become
\begin{eqnarray}
\bM \ddot{\bq} + \bh
&=& \bJ_c^{\rm T} \blambda_c,
\label{eq2.02a} \\
\bJ_c \dot{\bq} &=& {\bf 0}_{n \times 1},
\label{eq2.02b}
\end{eqnarray}
where
\begin{equation}
\bM 
=
{\small
\left[
\begin{array}{cccc}
\bM_1 & \bO & \bO & \bO \\
\bO & \bM_2 & \bO & \bO \\
\bO & \bO & \bM_3 & \bO \\
\bO & \bO & \bO & \bM_4
\end{array}
\right]
}, \ \
\bh = 
\frac{
\partial P_g
}{\partial \bq^{\rm T}} +
\frac{
\partial P_e
}{\partial \bq^{\rm T}} +
\frac{
\partial R
}{\partial \dot{\bq}^{\rm T}},
\end{equation}
and $\bM_i = {\rm diag} \left(
\begin{array}{ccc}
m_i & m_i & m_i a_i^2
\end{array}
\right)$. See \cite{CLAWAR2021} for the detail of $\bh$. The dimension
$n$ in Eq. (\ref{eq2.02b}) represents the number of velocity constraint
conditions, which is $6$ in the single-limb support (SLS) motion and $8$
in the double-limb support (DLS) motion.

In the following, we describe the detail of $\bJ_c$ in the case where
${\rm F}_{\rm 1A}$ is in contact with the floor as an example.

The velocity constraint conditions that
${\rm F}_{\rm 1A}$ is in contact with the floor without sliding are
specified as
\begin{eqnarray}
\dot{x}_{\rm f1a} &=& \dot{x}_1 - L_1 \dth_1 \cos \theta_1 = 0 ,
\label{eq2.03a}
\\
\dot{z}_{\rm f1a} &=& \dot{z}_1 + L_1 \dth_1 \sin \theta_1 = 0.
\label{eq2.03b}
\end{eqnarray}
In this paper, to form two cross-shaped frames, the following four
velocity constraint conditions that Limbs 1 and 3 and Limbs 2 and 4 are
connected to each other via the axis of rotation at each central point,
$G_i$, are added.
\begin{equation}
\dot{x}_1 = \dot{x}_3, \ \ \dot{z}_1 = \dot{z}_3, \ \   
\dot{x}_2 = \dot{x}_4, \ \ \dot{z}_2 = \dot{z}_4.
\label{eq2.04}
\end{equation}
By summarizing Eqs. (\ref{eq2.03a}), (\ref{eq2.03b}) and (\ref{eq2.04}),
we can obtain $\bJ_c$ as
\begin{eqnarray}
\!\!\!\! \bJ_c &=& \bJ_{\rm 1A} \nonumber \\
\!\!\!\!\ &=& 
\left[
\begin{array}{cccccccccccc}
1 & 0 & -L_1 \cos \theta_1 & 0 & 0 & 0 & 0 & 0 & 0 & 0 & 0 & 0 \\
0 & 1 & L_1 \sin \theta_1 & 0 & 0 & 0 & 0 & 0 & 0 & 0 & 0 & 0 \\
1 & 0 & 0 & 0 & 0 & 0 & -1 & 0 & 0 & 0 & 0 & 0 \\
0 & 1 & 0 & 0 & 0 & 0 & 0 & -1 & 0 & 0 & 0 & 0 \\
0 & 0 & 0 & 1 & 0 & 0 & 0 & 0 & 0 & -1 & 0 & 0 \\
0 & 0 & 0 & 0 & 1 & 0 & 0 & 0 & 0 & 0 & -1 & 0
\end{array}
\right].
\label{eq2.05}
\end{eqnarray}
As discussed in \cite{ICRA2010}, it is known that when this type of
walker starts passive-dynamic walking, the rigid frames gradually
synchronize with each other around the axis of rotation, and finally the
step length becomes too wide to continue stable walking. The models in
Fig. \ref{fig2.01} maintains a shape close to symmetry with sufficient
spring tension, this problem can be avoided.

When Limbs 1 and 3, Limbs 2 and 4 are constrained around the axis of
rotation and fixed in a state orthogonal to each other, it is necessary
to add the following two velocity constraint conditions.
\begin{equation}
\dth_1 = \dth_3, \ \ \dth_2 = \dth_4
\label{eq2.05a}
\end{equation}
By adding these equations, $\bJ_c$ should be
\begin{eqnarray}
\bJ_c &=& \bJ_{\rm 1A} \nonumber \\
&=& 
\left[
\begin{array}{cccccccccccc}
1 & 0 & -L_1 \cos \theta_1 & 0 & 0 & 0 & 0 & 0 & 0 & 0 & 0 & 0 \\
0 & 1 & L_1 \sin \theta_1 & 0 & 0 & 0 & 0 & 0 & 0 & 0 & 0 & 0 \\
1 & 0 & 0 & 0 & 0 & 0 & -1 & 0 & 0 & 0 & 0 & 0 \\
0 & 1 & 0 & 0 & 0 & 0 & 0 & -1 & 0 & 0 & 0 & 0 \\
0 & 0 & 0 & 1 & 0 & 0 & 0 & 0 & 0 & -1 & 0 & 0 \\
0 & 0 & 0 & 0 & 1 & 0 & 0 & 0 & 0 & 0 & -1 & 0 \\
0 & 0 & 1 & 0 & 0 & 0 & 0 & 0 & -1 & 0 & 0 & 0 \\
0 & 0 & 0 & 0 & 0 & 1 & 0 & 0 & 0 & 0 & 0 & -1
\end{array}
\right]
\label{eq2.05}
\end{eqnarray}
We call the VCRW model with the constraint of Eq. (\ref{eq2.05a}) VCRW1,
and that without it VCRW2.

The time derivative of Eq. (\ref{eq2.02b}) becomes
\begin{equation}
\bJ_c \ddot{\bq} + \dot{\bJ}_c \dot{\bq} = {\bf 0}_{n \times 1},
\label{eq2.06}
\end{equation}
and by solving this and Eqs. (\ref{eq2.02a}) for $\blambda_c \in
\mR^{n}$, we obtain
\begin{equation}
\blambda_c = \blambda_{\rm 1A} =
\left(
\bJ_{\rm 1A} \bM^{-1} \bJ_{\rm 1A}^{\rm T}
\right)^{-1}
\left(
\bJ_{\rm 1A} \bM^{-1}
\bh 
- \dot{\bJ}_{\rm 1A} \dot{\bq}
\right).
\label{eq2.07}
\end{equation}
Note that, however, the dimension of $\blambda_c$ changes to 6 or 8
depending on the number of rows in $\bJ_c$.

In the same procedure as above, the equations of $\bJ_c$ and
$\blambda_c$ should be changed according to the ground contact condition.

\subsection{Collision Equation}

Here, we describe the case where the stance foot switches from ${\rm
F}_{\rm 1A}$ to ${\rm F}_{\rm 2A}$ without the constraints of
Eq. (\ref{eq2.05a}) as an example. As in the model in \cite{CLAWAR2021},
in passive-dynamic walking of the models of Figs. \ref{fig2.01}(b) and
(c), non-instantaneous DLS motion continues in a short period after the
landing of the fore foot because the impact of one cross-shaped frame
does not affect the other. The collision equation that describes the
condition that ${\rm F}_{\rm 1A}$ is also in contact immediately after
${\rm F}_{\rm 2A}$ lands due to a complete inelastic collision, and the
holonomic constraint equation immediately after impact can be formulated
as follows.
\begin{eqnarray}
\bM \dot{\bq}^+ &=& \bM \dot{\bq}^- + \bJ_I^{\rm T} \blambda_I
\label{eq2.08a} \\
\bJ_I \dot{\bq}^+ &=& {\bf 0}_{8 \times 1}
\label{eq2.08b}
\end{eqnarray}
Here, the superscripts ``$-$'' and ``$+$'' denote immediately before and
immediately after impact. By solving Eqs. (\ref{eq2.08a}) and
(\ref{eq2.08b}), the velocity vector immediately after impact can be
obtained as
\begin{equation}
\dot{\bq}^+ = \left(
\bI_{12} - \bM^{-1} \bJ_I^{\rm T} \left(
\bJ_I \bM^{-1} \bJ_I^{\rm T}
\right)^{-1} \bJ_I
\right) \dot{\bq}^-.
\end{equation}

The detail of $\bJ_I$ for VCRW2 is described in the following. The
velocity constraint conditions where the rear foot ${\rm F}_{\rm 1A}$ is
in contact with the floor without sliding immediately after impact are
specified as
\begin{eqnarray}
\dot{x}_{\rm f1a}^{+} &=& \dot{x}_1^{+} - L_1 \dth_1^{+} \cos \theta_1 = 0 ,
\label{eq2.09a}
\\
\dot{z}_{\rm f1a}^{+} &=& \dot{z}_1^{+} + L_1 \dth_1^{+} \sin \theta_1 = 0.
\label{eq2.09b}
\end{eqnarray}
The same conditions for the fore foot ${\rm F}_{\rm 2A}$ are also
specified as
\begin{eqnarray}
\dot{x}_{\rm f2a}^{+} &=& \dot{x}_2^{+} - L_2 \dth_2^{+} \cos \theta_2 = 0 ,
\label{eq2.09c}
\\
\dot{z}_{\rm f2a}^{+} &=& \dot{z}_2^{+} + L_2 \dth_2^{+} \sin \theta_2 = 0.
\label{eq2.09d}
\end{eqnarray}
The following velocity constraint conditions must also hold at the same time.
\begin{equation}
\dot{x}_1^{+} = \dot{x}_3^{+}, \ \ \dot{z}_1^{+} = \dot{z}_3^{+}, \ \ 
\dot{x}_2^{+} = \dot{x}_4^{+}, \ \ \dot{z}_2^{+} = \dot{z}_4^{+}
\label{eq2.10a}
\end{equation}
By summarizing the eight conditions in Eqs. (\ref{eq2.09a}),
(\ref{eq2.09b}), (\ref{eq2.09c}), (\ref{eq2.09d}) and (\ref{eq2.10a}),
$\bJ_I$ is obtained as
\begin{eqnarray}
\bJ_I &=& \bJ_{\rm 1A2A} \nonumber \\
&=& \!
\left[
\begin{array}{cccccccccccc}
1 & 0 & -L_1 \cos \theta_1 & 0 & 0 & 0 & 0 & 0 & 0 & 0 & 0 & 0 \\
0 & 1 & L_1 \sin \theta_1 & 0 & 0 & 0 & 0 & 0 & 0 & 0 & 0 & 0 \\
0 & 0 & 0 & 1 & 0 & -L_2 \cos \theta_2 & 0 & 0 & 0 & 0 & 0 & 0 \\
0 & 0 & 0 & 0 & 1 & L_2 \sin \theta_2 & 0 & 0 & 0 & 0 & 0 & 0 \\
1 & 0 & 0 & 0 & 0 & 0 & -1 & 0 & 0 & 0 & 0 & 0 \\
0 & 1 & 0 & 0 & 0 & 0 & 0 & -1 & 0 & 0 & 0 & 0 \\
0 & 0 & 0 & 1 & 0 & 0 & 0 & 0 & 0 & -1 & 0 & 0 \\
0 & 0 & 0 & 0 & 1 & 0 & 0 & 0 & 0 & 0 & -1 & 0
\end{array}
\right]. \nonumber \\
\label{eq2.11}
\end{eqnarray}
Note that this $\bJ_I$ is also the same as $\bJ_c$ for the double-limb
support motion following this collision, and that the following
constraint conditions must be added to $\bJ_I$ for the inelastic
collision model of VCRW1.
\begin{equation}
\dth_1^+ = \dth_3^+, \ \ \dth_2^+ = \dth_4^+
\label{eq2.12}
\end{equation}

According to the stance foot condition, $\bJ_c$ transitions in the
following order while VCRW makes one rotation.
\begin{eqnarray*}
& & \bJ_{\rm 1A} \rightarrow
\bJ_{\rm 1A2A} \rightarrow
\bJ_{\rm 2A} \rightarrow
\bJ_{\rm 2A3A} \rightarrow
\bJ_{\rm 3A} \rightarrow
\bJ_{\rm 3A4A}
\\
& &
\rightarrow
\bJ_{\rm 4A} \rightarrow
\bJ_{\rm 4A1B} \rightarrow
\bJ_{\rm 1B} \rightarrow
\bJ_{\rm 1B2B} \rightarrow
\bJ_{\rm 2B} \rightarrow
\bJ_{\rm 2B3B} \\
& & \rightarrow
\bJ_{\rm 3B} \rightarrow
\bJ_{\rm 3B4B} \rightarrow
\bJ_{\rm 4B} \rightarrow
\bJ_{\rm 4B1A} \rightarrow
\bJ_{\rm 1A}
\end{eqnarray*}

As described in the next section, both VCRWs generate a stable limit
cycle gait consisting of single-limb support and double-limb support
phases. Since VCRW1 generates a passive-dynamic gait by using two rigid
leg frames, this is also a biped robot without swinging the swing leg.

\section{PASSIVE DYNAMIC WALKING}

\subsection{Typical Passive-dynamic Gaits}

\begin{figure}[!b]
\centering
\small
\footnotesize
\scalebox{0.67}{
\input{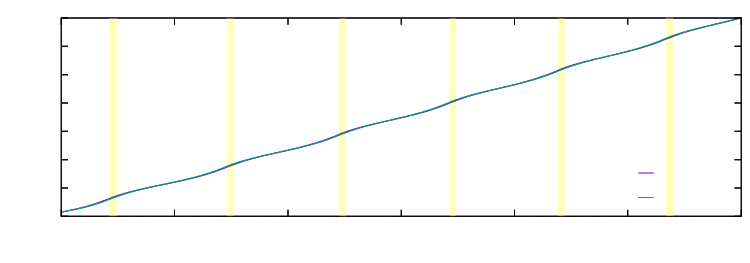}
}\\
(a) $x_i$\\
\scalebox{0.67}{
\input{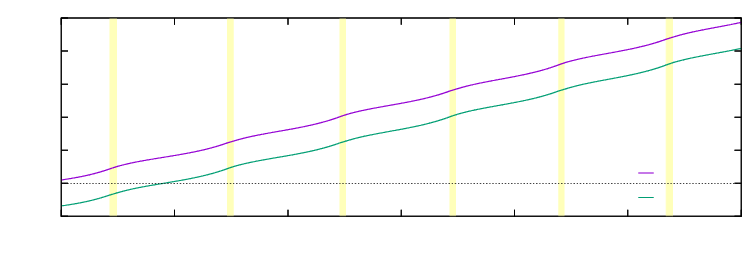}
}\\
(b) $\theta_i$\\
\scalebox{0.67}{
\input{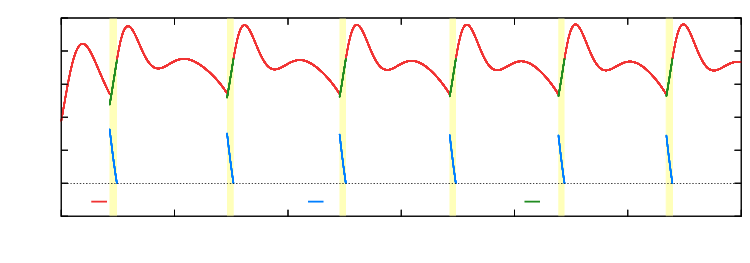}
}\\
(c) Vertical ground reaction force
\caption{Simulation results of passive dynamic walking of VCRW1}
\label{fig3.01}
\end{figure}

This section analyzes the basic passive-dynamic gait properties of VCRW1
and VCRW2. Fig. \ref{fig3.01} shows the simulation results of passive
dynamic walking of VCRW1 on a downhill where $\phi = 0.1$ [rad]. Here,
(a) is the time evolution of $x_i$, (b) that of $\theta_i$, and (c) that
of vertical ground reaction forces. The physical parameters were chosen
as the values listed in Table \ref{table01}. The DLS phases in each
figure were indicated by the light-yellow areas. VCRW1 started walking
from an appropriate initial state, and generated a stable, period-1
passive-dynamic gait. As seen in Figs. \ref{fig3.01}(a) and (b), the
relative central position and the relative angular position of the two
Greek-cross-shaped frames are hardly changed, and we can understand the
generated gait is almost the same as that of the normal eight-legged
RW. Note that, however, both $x_i$ and $\theta_i$ continue to increase
continuously because in this paper resetting the positional coordinates
immediately after every impact was not considered. From
Fig. \ref{fig3.01}(c), we can see that it generates a DLS motion in a
short period after the fore foot lands, and when the vertical ground
reaction force (VGRF) acting at the rear foot reaches zero while
monotonically decreasing, the rear leg leaves the floor and the next SLS
motion begins.

\begin{table}[!t]
\vspace*{3mm}
\caption{Physical parameters for VCRW1 and VCRW2}
\label{table01}
\centering
\renewcommand{\arraystretch}{1.2}
\small
\begin{tabular}{ccc} \hline
$m_1 = m_2 = m_3 = m_4$ & 1.0 & kg \\
$a_1 = a_2 = a_3 = a_4$ & 0.15 & m \\
$b_1 = b_2 = b_3 = b_4$ & 0.25 & m \\
$L_1 = L_2 = L_3 = L_4$ & 0.3 & m \\
$k$ & 200 & N/m \\
$c$ & 10 & N$\cdot$s/m \\
$L_0$ & 0.1 & m \\ \hline
\end{tabular}
\end{table}

\begin{figure}[!t]
\centering
\small
\footnotesize
\scalebox{0.67}{
\input{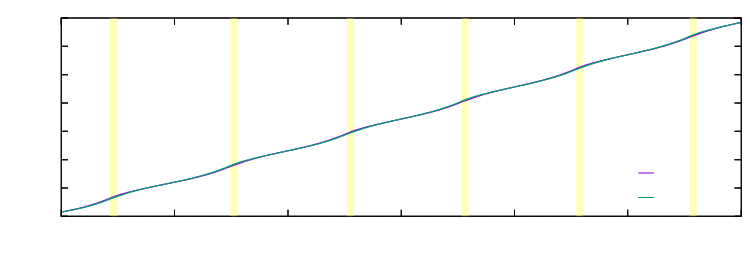}
}\\
(a) $x_i$\\
\scalebox{0.67}{
\input{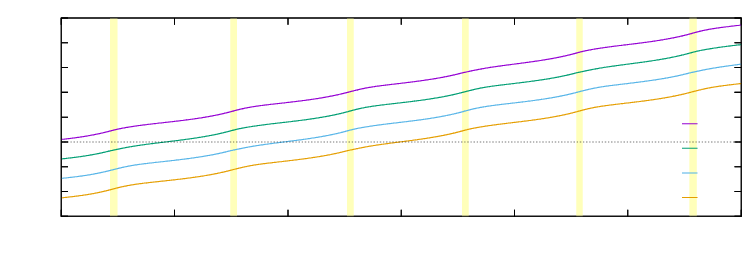}
}\\
(b) $\theta_i$\\
\scalebox{0.67}{
\input{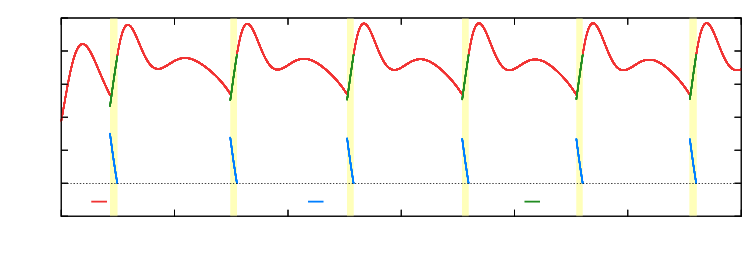}
}\\
(c) Vertical ground reaction force
\caption{Simulation results of passive dynamic walking of VCRW2}
\label{fig3.02}
\end{figure}

Fig. \ref{fig3.02} shows the simulation results of passive dynamic
walking of VCRW2 on a downhill where $\phi = 0.1$ [rad]. All the
simulation conditions were the same as above. We can see that the
generated gait is similar to that in Fig. \ref{fig3.01}, and that the
relative angles between the adjacent rigid frames are hardly changed
while maintaining a value close to $\pi/4$ [rad].

\subsection{Changes in Gait Properties with respect to Slope}

\begin{figure*}[!t]
\vspace*{2mm}
\hspace*{-6mm}
\begin{minipage}{0.32\linewidth}
\scalebox{0.60}{
\input{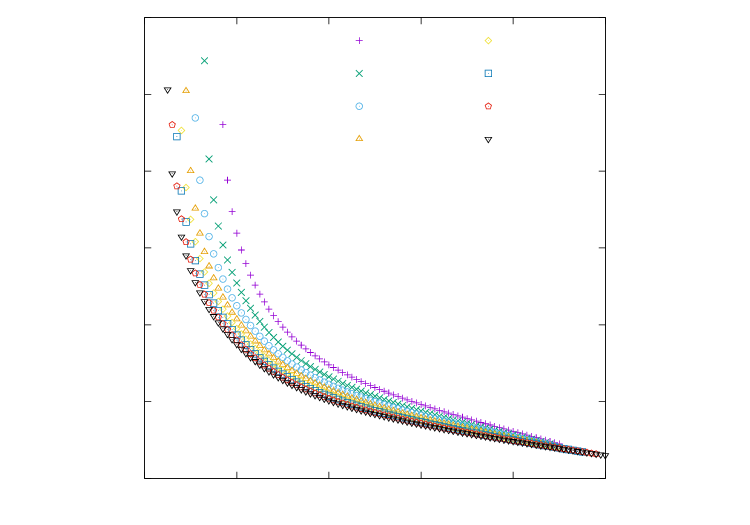}
}
\end{minipage}
\begin{minipage}{0.32\linewidth}
\scalebox{0.60}{
\input{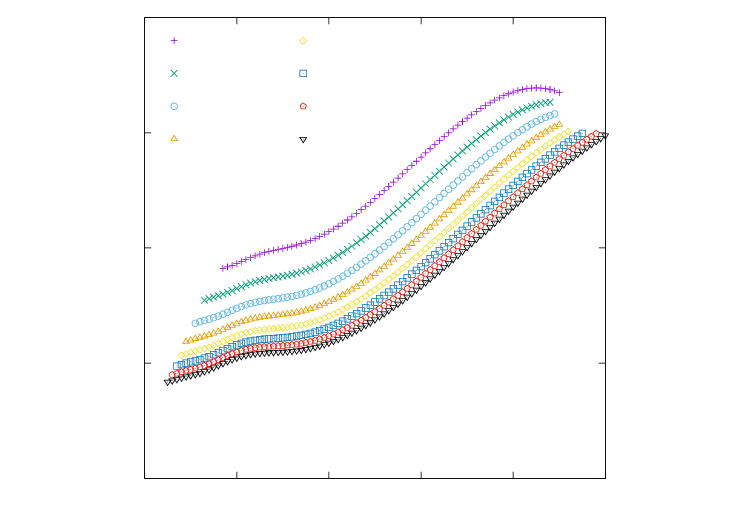}
}
\end{minipage}
\begin{minipage}{0.32\linewidth}
\scalebox{0.60}{
\input{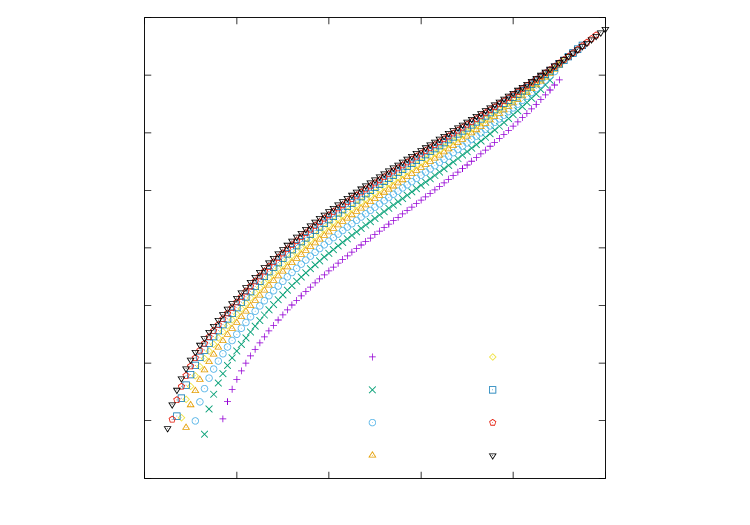}
}
\end{minipage}
\bigskip
\caption{Gait descriptors versus slope for eight values of $b_i$ in
 passive dynamic walking of VCRW1}
\label{fig3.03}
\end{figure*}

Fig. \ref{fig3.03} shows the gait descriptors versus slope in passive
dynamic walking of VCRW1 for eight values of $b_i$. Here, (a) is the
step period, (b) the step length, and (c) the walking speed. The
physical parameters except $b_i$ were chosen as the same values listed
in Table \ref{table01} again. By the effect of the dampers, all the
generated gaits converged to a period-1 limit cycle. We then plotted the
mean value of the data for 20 steps after 100 seconds had passed since
the start of walking. We can see that the step period decreases 
monotonically, whereas the step length increases almost monotonically as
the slope increases. Since the walking speed, the value obtained by
dividing the step length by the step period, increases monotonically as
the slope increases, we can conclude that the change in the step period
has a greater effect on this than the step length. VCRW1 has the same
gait properties as a normal RW, except that the step length increases a
little as the slope increases.

\begin{figure*}[!t]
\hspace*{-6mm}
\begin{minipage}{0.32\linewidth}
\scalebox{0.60}{
\input{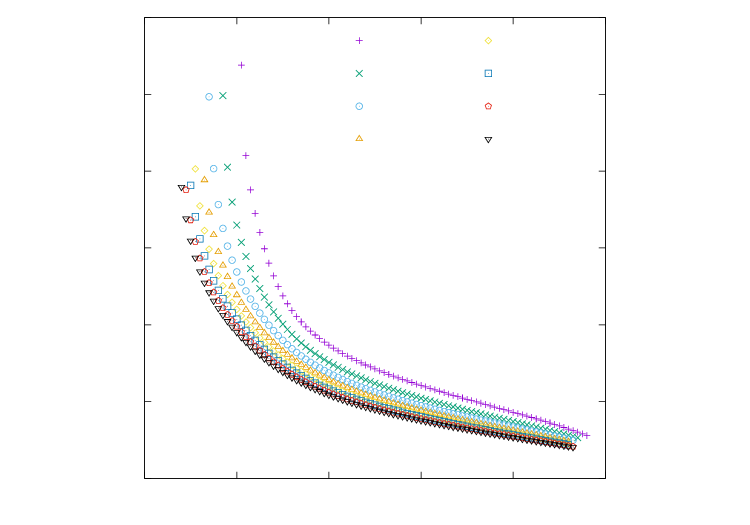}
}
\end{minipage}
\begin{minipage}{0.32\linewidth}
\scalebox{0.60}{
\input{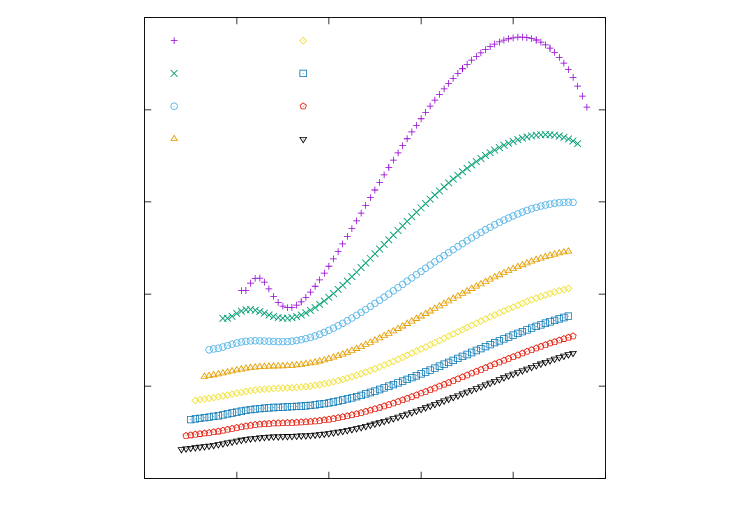}
}
\end{minipage}
\begin{minipage}{0.32\linewidth}
\scalebox{0.60}{
\input{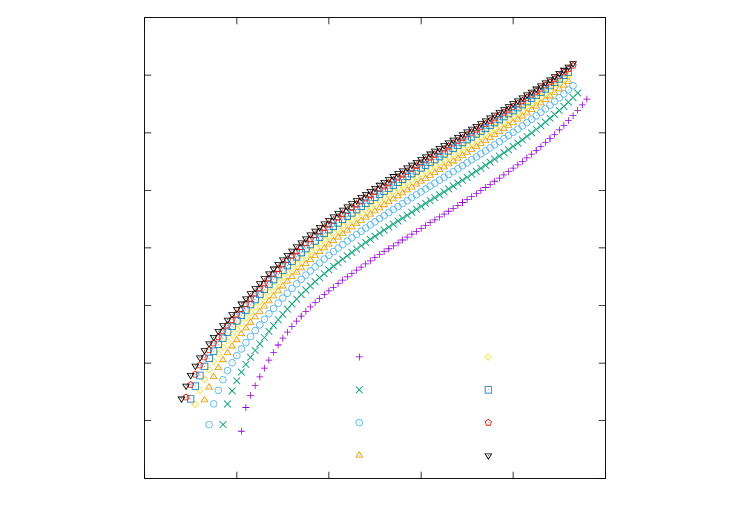}
}
\end{minipage}
\bigskip
\caption{Gait descriptors versus slope for eight values of $b_i$ in
 passive dynamic walking of VCRW2}
\label{fig3.04}
\end{figure*}

Fig. \ref{fig3.04} shows the gait descriptors versus slope in passive
dynamic walking of VCRW2 for eight values of $b_i$. All the calculation
procedures were the same as above. We can see that the step period and
walking speed show the similar range and change tendency as in
VCRW1. Since VCRW2 does not have the rotational constraint around the
central axis, the impact posture can be changed more flexibly, and the
changing tendency of the step length is more complicated. In both VCRWs,
the smaller $b_i$, the longer the step length, and the more complicated
the changing tendency with respect to the slope. This is because if the
natural length of the spring, $L_0$, is constant, the average spring
tension becomes weaker as $b_i$ becomes smaller.

\section{ACTIVATION FOR ASSIST WALKING}

This section discusses an extension to level-ground walking via
activation of one of the cross-shaped frames in VCRW2. An upper-body
link was added to one cross-shaped frames, and leave the other
unchanged; we call it VCRW3. As illustrated in Fig. \ref{fig2.01}(c),
let the length, the mass, the inertia moment and the angular position of
the upper-body link be $L_5$, $m_5$, $I_5$ and $\theta_5$. It is assumed
that the control torque, $u$, can be exerted clockwise from the upper
body to Limb 1. VCRW3 becomes 15-DOF by adding the upper-body link and
the system equations have been changed accordingly, but the details are
omitted here.

In the following, we outline the control system design. Let $\theta_5$
be the control output. We then consider an input-output linearization as
follows.
\begin{eqnarray}
\ddot{\theta}_5 &=& A u - B \\
u &=& A^{-1} \left( v + B \right) \\
v &=& - K_D \dth_5 - K_P \left( \theta_{5{\rm d}} - \theta_5 \right)
\end{eqnarray}
Here, $A$ and $B$ are the scalar functions of the state variables, and
$v$ is the acceleration command signal. In this paper, the target angle
of $\theta_5$ was chosen as $\theta_{5{\rm d}} = 0.3$ [rad]. As a result
of following the upper-body angle to this constant value, a clockwise
rotation torque is applied to Limb 1, and the cross-shaped frame with
the upper body continues to rotate so as to move forward.

It is difficult for both the activated four-legged RW and the passive
one to walk on their own, but by combining them, it can walk
smoothly. Therefore, this can be regarded as walking in cooperation with
each other, or walking while the active four-legged RW supports the
passive one. The authors consider that VCRW3 is an abstraction of a
support system such as exoskeleton suits or crutches \cite{Zhou,Lov,Matsuura}.

\begin{figure}[!t]
\centering
\vspace*{1mm}
\small
\footnotesize
\scalebox{0.67}{
\input{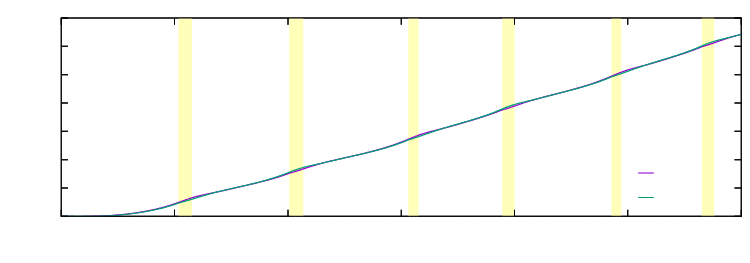}
}\\
(a) $x_i$ \\
\scalebox{0.67}{
\input{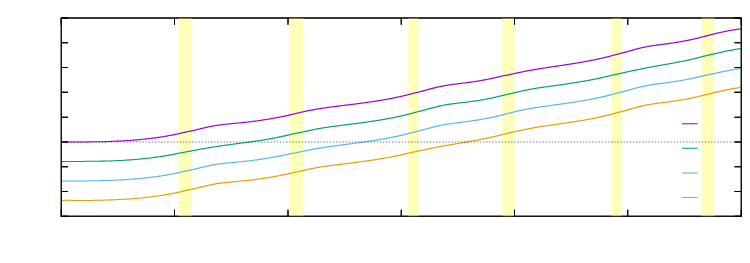}
}\\
(b) $\theta_i$ \\
\scalebox{0.67}{
\input{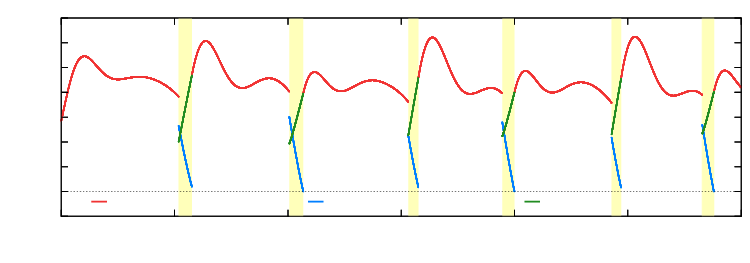}
}\\
(c) Vertical ground reaction forces
\caption{Simulation results of limit cycle walking on level ground of VCRW3}
\label{fig4.01}
\vspace*{-2.2mm}
\end{figure}

\begin{figure}[!t]
\centering
\vspace*{1mm}
\small
\footnotesize
\scalebox{0.67}{
\input{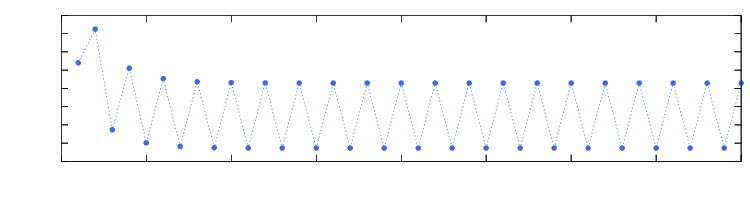}
}\\
(a) Step period\\
\scalebox{0.67}{
\input{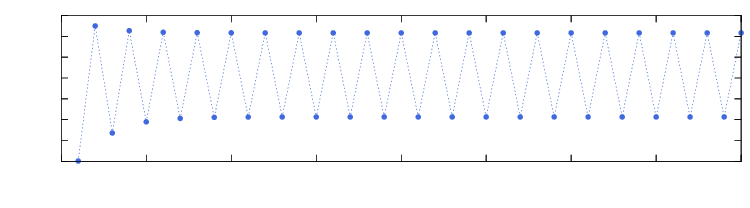}
}\\
(b) Step length\\
\scalebox{0.67}{
\input{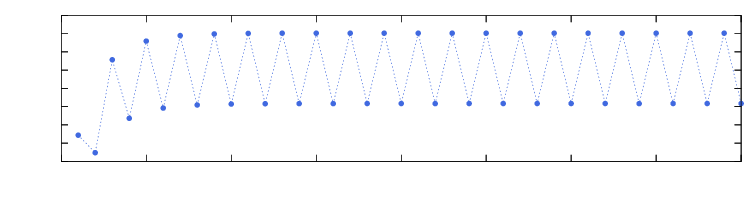}
}\\
(c) Walking speed\\
\scalebox{0.67}{
\input{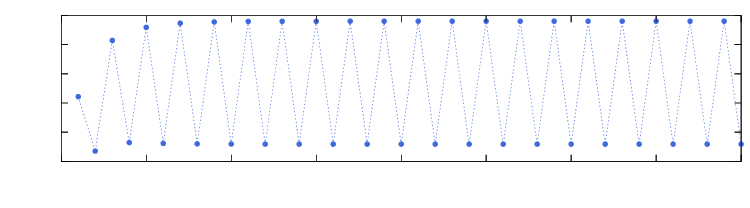}
}\\
(d) Specific resistance\\
\caption{Evolution of gait descriptors in limit cycle walking on level ground}
\label{fig4.02}
\vspace*{-2mm}
\end{figure}

Fig. \ref{fig4.01} shows the simulation results of active-dynamic
walking on level ground, and Fig. \ref{fig4.02} shows the evolution of
the gait descriptors. VCRW3 started walking from the same initial state
as in the previous section. The physical parameters of the upper body
were chosen as $L_5 = 0.3$ [m], $m_5 = 1.0$ [kg] and $I_5 = 0.0225$
[N$\cdot$m$^2$], respectively. The PD gains were also chosen as $K_D =
20$ [s$^{-1}$] and $K_P = 100$ [s$^{-2}$]. 
The other parameters were chosen as the values listed in
Table \ref{table01} again. From the results, we can see that a stable
period-2 limit-cycle gait has been generated because this is composed of
two different walkers. This system can be regarded as a walking support
system in which the activated cross-shaped frame supports the passive
one that cannot move at all by itself, while cooperating to generate a
stable gait on the horizontal plane. 

The energy efficiency of the generate gait can be evaluated in terms of
specific resistance (SR). As seen in Fig. \ref{fig4.02}(d), the average
value of SR in the generate gait was about $0.1054$, and this is very
small in general limit cycle gaits with an upper body. By setting the
target upper-body angle to a larger value, the walking speed can be
increased, but there will be a trade-off between the energy efficiency
and the walking speed.

\begin{figure}[!t]
\centering
\small
\vspace*{-9mm}
\scalebox{0.67}{
\input{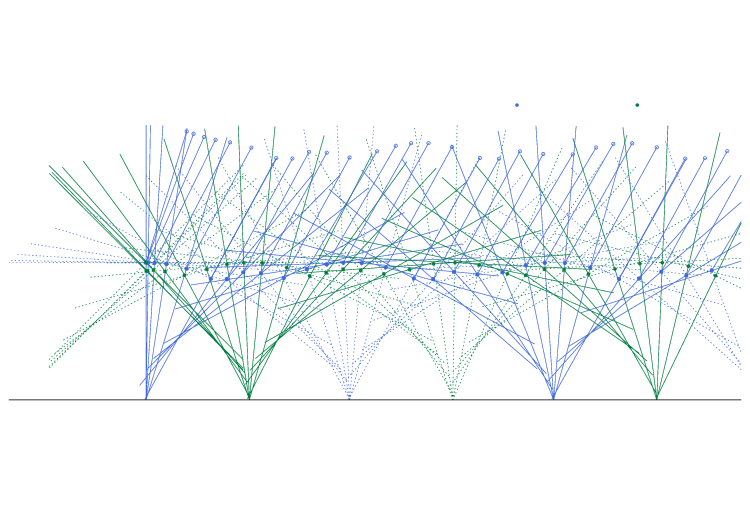}
}
\vspace*{-18mm}
\caption{Stick diagram for generated level gait in Fig. \ref{fig3.01}}
\label{fig4.03}
\vspace*{-1.5mm}
\end{figure}

Fig. \ref{fig4.03} plots the stick diagram of the generated level gait
in Fig. \ref{fig4.01} for the first several steps. The positions of the
central axes of the two cross-shaped frames were indicated by filled
circles. Compared to the normal RW, we can understand that the COM
position of the entire VCRW draws a flatter orbit if the upper body is
not taken into account. By dividing the eight-legged RW body into two
four-legged ones and utilizing the effect of the viscoelastic elements
attached to the whole body, an environment-adaptive, crush-worthy
limit-cycle gait has been generated.


\section{CONCLUSION AND FUTURE WORK}

In this paper, we proposed the VCRW robots that combined the advantages
of the conventional RWs and the flexibility due to the tensegrity
structure. The proposed walkers not only maintains the properties of
stable passive-dynamic and limit-cycle walking, but also reduces adverse
effects at impact on the gait stability by distributing the external
force to each spring, so it can be expected that highly adaptable
walking can be achieved. The generated gait properties were analyzed and
the relationship between its own structure and gait was mainly
investigated through different models. Furthermore, based on this, the
upper-body link was added for activation, and not only level walking can
be achieved, but also the upper-body balance can be maintained according
to the control method, which provides a theoretical basis or an
abstraction for walking assist devices. The authors consider that the
proposed VCRW3 model is more crash-worthy and adaptable than
conventional walking assist devices.

The VCRW models proposed in this paper have intermediate physical
properties between a normal eight-legged RW and a tensegrity robot model
in which each rigid frame can move independently only by the effect of
viscoelastic elements and gravity \cite{CLAWAR2021}. In addition to
comparing and examining the gait efficiency of these models in more
detail, we would like to investigate not only walking but also
evaluation of impact resistance when falling. 

\begin{thebibliography}{10}
\bibitem{TRO}
C. Paul, F. J. Valero-Cuevas and H. Lipson, ``Design and control of
	tensegrity robots for locomotion,'' {\it IEEE Trans. on
	Robotics}, Vol. 22, Iss. 5, pp. 944--957, 2006.
\bibitem{ICRA2015}
A. P. Sabelhaus, J. Bruce, K. Caluwaerts, P. Manovi, R. F. Firoozi,
	S. Dobi, A. M. Agogino and V. SunSpiral, ``System design and
	locomotion of SUPERball, an untethered tensegrity robot,'' {\it
	Proc. of the IEEE Int. Conf. on Robotics and Automation},
	pp. 2867--2873, 2015.
\bibitem{Hirai}
W. Du, S. Ma, B. Li, M. Wang and S. Hirai, ``Force analytic method for
	rolling gaits of tensegrity robots,'' {\it IEEE/ASME Trans. on
	Mechatronics}, Vol. 21, Iss. 5, pp. 2249--2259, 2016.
\bibitem{Coleman}
M. J. Coleman, ``Dynamics and stability of a rimless spoked wheel: a
	simple 2D system with impacts,'' {\it Dynamical Systems},
	Vol. 25, No. 2, pp. 215--238, 2010.
\bibitem{McGeer}
T. McGeer, ``Passive dynamic walking,'' {\it The Int. J. of Robotics
	Research}, Vol. 9, Iss. 2, pp. 62--82, 1990.
\bibitem{CLAWAR2021}
F. Asano, Y. Zheng and L. Li, ``Modeling and motion analysis of planar
	passive-dynamic walker with tensegrity structure formed by four
	limbs and eight viscoelastic elements,'' {\it Proc. of the 24th
	Int. Conf. Series on Climbing and Walking Robots}, pp. 242--254,
	2021.
\bibitem{Zheng2}
Y. Zheng, F. Asano, L. Li and C. Yan, ``Analysis of passive dynamic gait
	of tensegrity robot,'' {\it Proc. of the 24th Int. Conf. Series
	on Climbing and Walking Robots}, pp. 274--285, 2021.
\bibitem{Zheng3}
Y. Zheng, L. Li, F. Asano, C. Yan, X. Zhao and H. Chen, ``Modeling and
	analysis of tensegrity robot for passive dynamic walking,'' {\it
	Proc. of the IEEE/RSJ Int. Conf. on Intelligent Robots and
	Systems}, pp. 2456--2461, 2021.
\bibitem{Zhao}
H. Zhao, F. Asano and L. Li, ``Indirectly controlled combined rimless
	wheel that consists of eight- and ten-legged wheels via
	entrainment effect,'' {\it Proc. of the 3rd Int. Symp. on Swarm
	Behavior and Bio-Inspired Robotics}, pp. 250--253, 2019.
\bibitem{MUBO}
F. Asano and I. Tokuda, ``Indirectly controlled limit cycle walking of
	combined rimless wheel based on entrainment to active wobbling
	motion,'' {\it Multibody System Dynamics}, Vol. 34, Iss. 2,
	pp. 191--210, 2015.
\bibitem{MIT}
D. J. Gonzalez and H. H. Asada, ``Passive quadrupedal gait
	synchronization for extra robotic legs using a dynamically
	coupled double rimless wheel model,'' {\it Proc. of the IEEE
	Int. Conf. on Robotics and Automation}, pp. 3451--3457, 2020.
\bibitem{ICRA2010}
F. Asano, ``Simulation and experimental studies on passive-dynamic
	walker that consists of two identical crossed frames,'' {\it
	Proc. of the IEEE Int. Conf. on Robotics and Automation},
	pp. 1703--1708, 2010.
\bibitem{Zhou}
J. Zhou, S. Yang and Q. Xue, ``Lower limb rehabilitation exoskeleton
	robot: A review,'' {\it Advanced in Mechanical Engineering},
	Vol. 13, Iss. 4, pp. 1--17, 2021.
\bibitem{Lov}
Z. Lovrenovic and M. Doumit, ``Development and testing of a passive
	walking assist exoskeleton,'' {\it Biocybernetics and Biomedical
	Engineering}, Vol. 39, Iss. 4, pp. 992--1004, 2019.
\bibitem{Matsuura}
D. Matsuura, R. Funato, M. Ogata, M. Higuchi and Y. Takeda, ``Efficiency
	improvement of walking assist machine using crutches based on
	gait-feasible region analysis, {\it Mechanism and Machine
	Theory}, Vol. 84, pp. 126--133, 2015.
\end{thebibliography}
\end{document}